\documentclass{ieeeaccess}
\usepackage{cite}
\usepackage{amsmath,amssymb,amsfonts}
\usepackage{algorithmic}
\usepackage{graphicx}
\usepackage{textcomp}

\usepackage{bm}
\makeatletter
\AtBeginDocument{\DeclareMathVersion{bold}
\SetSymbolFont{operators}{bold}{T1}{times}{b}{n}
\SetSymbolFont{NewLetters}{bold}{T1}{times}{b}{it}
\SetMathAlphabet{\mathrm}{bold}{T1}{times}{b}{n}
\SetMathAlphabet{\mathit}{bold}{T1}{times}{b}{it}
\SetMathAlphabet{\mathbf}{bold}{T1}{times}{b}{n}
\SetMathAlphabet{\mathtt}{bold}{OT1}{pcr}{b}{n}
\SetSymbolFont{symbols}{bold}{OMS}{cmsy}{b}{n}
\renewcommand\boldmath{\@nomath\boldmath\mathversion{bold}}}
\makeatother

\def\BibTeX{{\rm B\kern-.05em{\sc i\kern-.025em b}\kern-.08em
    T\kern-.1667em\lower.7ex\hbox{E}\kern-.125emX}}

\begin{document}
\history{Date of publication xxxx 00, 0000, date of current version xxxx 00, 0000.}
\doi{10.1109/ACCESS.2024.0429000}

\title{Robust Classification of Oral Cancer with Limited Training Data}

%\authorrefmark{1}, \authorrefmark{2}, \authorrefmark{3}
% \IEEEmembership{Fellow, IEEE}, %\IEEEmembership{Member, IEEE}}
\author{\uppercase{Akshay Bhagwan Sonawane},
\uppercase{Lena D. Swamikannan}, 
and \uppercase{Lakshman Tamil}}

\address{The authors are with Quality of Life Technology Laboratory, The University of Texas at Dallas, Richardson, TX 75080, USA. (e-mail:akshaybhagwan.sonawane@utdallas.edu, lena.duraisamyswamikannan@utdallas.edu, laxman@utdallas.edu).}
% \address[2]{Department of Physics, Colorado State University, Fort Collins,
% CO 80523 USA (e-mail: author@lamar.colostate.edu)}
% \address[3]{Electrical Engineering Department, University of Colorado, Boulder, CO
% 80309 USA}
% \tfootnote{This paragraph of the first footnote will contain support
% information, including sponsor and financial support acknowledgment. For
% example, ``This work was supported in part by the U.S. Department of
% Commerce under Grant BS123456.''}

\markboth
{Akshay \headeretal: Robust Classification of Oral Cancer with Limited Training Data}
{Akshay \headeretal: Robust Classification of Oral Cancer with Limited Training Data}

\corresp{Corresponding author: Akshay B. Sonawane (e-mail: akshaybhagwan.sonawane@utdallas.edu).}

\begin{abstract}
Oral cancer ranks among the most prevalent cancers globally, with a particularly high mortality rate in regions lacking adequate healthcare access. Early diagnosis is crucial for reducing mortality; however, challenges persist due to limited oral health programs, inadequate infrastructure, and a shortage of healthcare practitioners. Conventional deep learning models, while promising, often rely on point estimates, leading to overconfidence and reduced reliability. Critically, these models require large datasets to mitigate overfitting and ensure generalizability—an unrealistic demand in settings with limited training data. To address these issues, we propose a hybrid model that combines a convolutional neural network (CNN) with Bayesian deep learning for oral cancer classification using small training sets. This approach employs variational inference to enhance reliability through uncertainty quantification. The model was trained on photographic color images captured by smartphones and evaluated on three distinct test datasets. The proposed method achieved 94\% accuracy on a test dataset with a distribution similar to that of the training data, comparable to traditional CNN performance. Notably, for real-world photographic image data—despite limitations and variations differing from the training dataset—the proposed model demonstrated superior generalizability, achieving 88\% accuracy on diverse datasets compared to 72.94\% for traditional CNNs, even with a smaller dataset. Confidence analysis revealed that the model exhibits low uncertainty (high confidence) for correctly classified samples and high uncertainty (low confidence) for misclassified samples. These results underscore the effectiveness of Bayesian inference in data-scarce environments in enhancing early oral cancer diagnosis by improving model reliability and generalizability.
\end{abstract}


\titlepgskip=-21pt

\maketitle

\section{Introduction}
\label{sec:introduction}
\PARstart{O}{ral} diseases are significant contributors to major health issues worldwide, alongside other noncommunicable diseases. According to the World Health Organization (WHO), nearly 3.5 billion people globally are affected by oral diseases, with three out of four affected individuals residing in low- and middle-income countries \cite{b1}. Although many oral diseases are preventable, it is crucial to recognize the more severe risks, such as oral cancer, which ranks as the 13th most common cancer worldwide, with high mortality and morbidity rates. Oral cancer encompasses cancer lesions occurring on the lips, various parts of the mouth, and the oropharynx. In 2020 alone, an estimated 377,713 new cases and 177,757 deaths from oral cancer were reported, with global age-standardized mortality rates (ASMR) of 2.8 per 100,000 men and 1 per 100,000 women \cite{b1,b2}. The major risk factors associated with oral cancer include high tobacco smoking rates, excessive alcohol consumption in Western countries, and the use of betel quid and tobacco chewing in South and Southeast Asia \cite{b1}. Early diagnosis is crucial for reducing the mortality rate of oral cancer. However, limited promotive and preventive oral health programs in universal healthcare have hindered access for the general population. In addition, the lack of healthcare infrastructure and practitioners poses challenges in reaching high-risk populations. Here, we present an opportunity to leverage the potential of deep learning to combat oral cancer.

Deep learning techniques, which are renowned for their scalability, adaptability, and high predictive power, have demonstrated remarkable success across various domains, including Natural Language Processing, Computer Vision, and Time Series analysis. In the healthcare sector, particularly in computer vision, deep convolutional neural networks have been effectively used for diagnosing skin, breast, lung, and oral cancers. Because of their outstanding performance, many of these networks are being considered for integration into automatic diagnostic systems \cite{b3}. However, the direct application of deep neural networks without reliability and explainability can be detrimental, especially in high-risk sectors like healthcare. Deep neural networks, which rely on point estimates for their weights for inference, can exhibit overconfidence in their outputs and fail to adequately consider the uncertainties inherent in the real world. Consequently, they cannot mimic the diagnostic abilities of oral care practitioners who routinely navigate diagnostic uncertainties.

To address this challenge, the Bayesian framework is proposed as a promising solution. Bayesian methods have garnered considerable interest in deep learning because they offer an alternative to classical methods by providing probability distributions rather than point estimations for inputs. By imposing Bayesian probability theory, reliability can be embedded into deep neural networks. Bayesian deep learning (BDL) achieves this by integrating prior distributions over weights and deriving posterior distributions, thereby providing predictions with associated uncertainties. Unlike traditional deep learning models that yield deterministic outputs, BDL estimates uncertainty using a probability density over outcomes \cite{b3}. Furthermore, despite the widespread acceptance of traditional deep neural networks, they are prone to overfitting, which compromises their generalizability and performance on unseen data. This is because deep neural networks require significant amounts of data, and gathering large datasets in healthcare environments presents another major challenge. However, BDL has proven to perform better, particularly when data availability is limited, by generalizing more effectively and mitigating overfitting through predictions based on probability distributions over weights.

In this study, we propose a Bayesian neural network model for the classification of oral cancer based on color images of the mouth. Given the limited availability of annotated photo images of oral cancer, this study focuses on developing a model that can perform reliably under data-scarce conditions. The proposed model employs a Bayesian framework with variational inference to deliver predictions and associated uncertainties. Our Bayesian neural network performs comparably to traditional deep neural networks on our proprietary test dataset. In addition, the proposed Bayesian model demonstrated superior generalization across the Kaggle dataset \cite{b18}, which exhibits different distributions than our proprietary dataset. This highlights the strength of our approach in achieving robust performance and generalization despite the data limitation. Despite the potential of BDL, it remains underused by the research community in the healthcare domain. The findings of this study can serve as a guide for future work to enhance the incorporation of BDL in various healthcare areas.

The remainder of this paper is organized as follows: Section II discusses the literature review related to this research. Section III explains the research methods used in this study. Section IV introduces the datasets used in this study. In Section V, we examine the results and discuss the performance of the proposed BDL model. Section VI reports the conclusions and future work.

\section{Related Work}
Many studies have been performed to automate the detection and classification of oral cancer using different imaging techniques, such as histopathological, hyperspectral, autofluorescence, white light, and color imaging. Machine learning techniques including deep learning, and feature extraction techniques are primarily used to develop automatic diagnostic systems.

In the imaging field prior to the rise of deep learning, researchers employed classical machine learning techniques along with feature extraction methods. Krishnan \emph{et al.} \cite{b5} used histopathological images and developed a hybrid feature extraction approach that combined higher-order spectra, local binary patterns, and low texture energy to classify normal, oral submucous fibrosis without dysplasia, and oral submucous fibrosis with dysplasia. Their method achieved 95.7\% accuracy using a Fuzzy classifier. Wang \emph{et al.} \cite{b6} incorporated color normalization, automatic sampling, and principal component analysis and proposed a color-based approach for tumor tissue classification. More recently, Rahman \emph{et al.} \cite{b7} proposed a method to identify oral squamous cell carcinoma using shape, texture, and color features extracted from whole image strips. They achieved 100\% accuracy using SVM and Logistic Regression for textural features and 100\% accuracy using Linear Discriminant Analysis for color features.

Building on advances in artificial intelligence, deep learning approaches have rapidly gained traction for oral cancer detection and diagnosis. Uthoff \emph{et al.} \cite{b8} proposed the use of convolutional neural networks (CNNs) in conjunction with smartphone-based devices employing autofluorescence and white-light imaging, demonstrating promising results in classifying oral lesions. Furthermore, studies like Refs.\cite{b9} and \cite{b10} have achieved high accuracy in distinguishing cancerous from normal tissues by applying deep learning algorithms to clinical oral photographs. These machine learning models have demonstrated accuracy ranging from 85\% to 100\%, with sensitivity and specificity exceeding 80\% \cite{b9} -\cite{b11}. The application of deep learning in histopathological images has also shown potential for diagnosing, classifying, and predicting oral cancer outcomes \cite{b11}. These advancements suggest that AI-based tools can significantly enhance the accuracy and accessibility of oral cancer screening, particularly in low-resource settings\cite{b8} - \cite{b10}.

Despite the improved performance of the ML models, the above studies lack the reliability of the model predictions, which is important in the healthcare sector. To address this issue, many researchers have focused on the use of Bayesian methods in machine learning models. Various studies have used Bayesian deep learning approaches to measure uncertainties to improve the reliability and accuracy of medical image classification tasks. Thiagarajan \emph{et al.} \cite{b12} comprehensively compared Bayesian and traditional CNNs using a variational inference method \cite{b14}. They demonstrated that Bayesian CNNs outperformed traditional CNNs in histopathological image classification by reducing both false positives and negatives, while using fewer parameters. Furthermore, Thiagarajan \emph{et al.} \cite{b12} and Raczkowski \emph{et al.} \cite{b13} demonstrated that we can improve overall classification accuracy, enable active learning, and detect mislabeled samples through the uncertainty measures provided by Bayesian models. Additionally, Taher \emph{et al.} \cite{b15} used Bayesian classification methods and demonstrated the potential of sputum image analysis for early lung cancer diagnosis.

Bayesian deep learning is gaining popularity, and it has been used to diagnose breast \cite{b12}, colorectal \cite{b13}, and lung \cite{b15} cancer. However, very few studies, to the best of authors' knowledge, have demonstrated the advantage of Bayesian deep neural networks for diagnosing oral cancer using color images and describing their generalization capabilities. One such study was by Song \emph{et al.} \cite{b16}, who used the Monte Carlo (MC) dropout \cite{b17} method and developed a Bayesian deep network capable of estimating uncertainty for assessing the reliability of oral cancer image classification. However, models developed using MC dropout methods are generally overconfident regarding their predictions. In addition, the MC dropout methods are not truly Bayesian in the sense that they loosely approximate the Bayesian inference. Moreover, to best of authors' knowledge, most studies on oral cancer have developed diagnostic systems that use histopathological cellular or color images using intraoral screening devices. Thus, the use of true Bayesian methods should be explored to improve the reliability of color images in oral cancer screening.

Therefore, in this study, we proposed a hybrid Bayesian neural network model that uses variational inference to diagnose oral cancer. We used photographic oral images captured with a smartphone to diagnose the disease and compared the performance of our model with that of a traditional CNN. In addition, we compared the generalization capabilities of the Bayesian and traditional models on datasets of different distributions. Finally, we evaluated the uncertainty quantification of the proposed model on different test datasets.

\section{Methodology}
Bayesian neural networks have recently attracted considerable interest as an alternative to classical neural network models. Although classical deep learning models perform extremely well on various healthcare problems, they are generally incapable of addressing uncertainty. The point estimation of weights for the classical modeling setup predicts the deterministic output, which can sometimes be overconfident. Bayesian deep learning methods implicitly allow the quantification of uncertainties in both estimated parameters and predictions. This is achieved by estimating the probability distribution over the weights of the deep learning models. The formulation of the Bayesian perspective for the neural network is summarized using the Bayes theorem, as shown in \eqref{eq1}.
\begin{equation}
p(\theta | D) = \frac{p(\theta)p(D|\theta)}{\int_{\theta}p(D|\theta)p(\theta)\,d\theta} \label{eq1}
\end{equation}

The primary goal of the Bayesian neural network is to estimate the posterior distribution, which is denoted by $p(\theta|D)$. Based on the Bayes theorem, to find the posterior distribution, we consider the prior probability $p(\theta)$, which is the prior knowledge about the parameter before observing the data. The parameter $\theta$ is the k-dimensional vector of the weights of the postulated model. The posterior distribution is then estimated by mixing prior knowledge with evidence supported by the model's likelihood \cite{b4}. Here, the model's likelihood is denoted by $p(D|\theta)$. Because of this mixture, Bayesian inference becomes useful in applications like healthcare, where collecting large datasets is challenging, as it can generalize well even with small datasets. The $\int_{\theta}p(D|\theta)p(\theta)\,d\theta$ is the evidence, also known as the marginal likelihood, obtained after collecting the data. As a normalization constant, it retrieves the proper probability distribution for $p(\theta|D)$. Although estimating the marginal likelihood is important for capturing the true posterior distribution, it is often quite complicated to compute.

The nontrivial computation of marginal likelihood in Bayesian inference is often intractable and unknown \cite{b4}. Except for a limited class of prior distributions whose parametric forms are similar to likelihoods, the evidence computation is analytically tractable. To address this issue, various Monte Carlo (MC) methods have been investigated; however, they are challenging and infeasible when dealing with high-dimensional parametric models like neural networks. This is due to their intrinsic need for high-dimensional sampling techniques. We use variational inference (VI) algorithmic methods.

Variational inference is an approximation method in which we attempt to approximate the posterior distribution. In this approach, we use a prior from a tractable distribution, such as an exponential family, and impose it over the weights of the neural network. We then parameterize the components of the prior distribution, for example, in the case of a multivariate Gaussian distribution $\zeta = [\mu,\sigma]$, and solve the optimization problem by minimizing the Kullback-Leibler (KL) divergence from the variational distribution $q(\theta|\zeta)$ to the prior distribution $p(\theta)$. The main task here is to find the optimized variational parameters such that the final variational posterior approximates the real unknown posterior. In the neural network training, the optimization problem is defined as in \eqref{eq2}.
\begin{equation}
\begin{split}
\mathcal{L}(\mathcal{D}_{tr},\zeta) = argmin_\zeta \sum_{(x_i, y_i) \epsilon \mathcal{D}_{tr}} \mathcal{L}(F_{\theta}(x_i), y_i)\\
+ KL[q(\theta|\zeta) \parallel p(\theta)]
\end{split}
\label{eq2}
\end{equation}

Here, $\mathcal{L}(F_{\theta}(x_i), y_i)$ is the likelihood of the neural network model $ F_{\theta}$. $x_i$ and $y_i$ are instances of the data from dataset $\mathcal{D}_{tr}$. Once we find the optimized parameters of the variational posterior distribution during the test time, we can approximate the real posterior distribution for an instance, as shown in \eqref{eq3}. The final prediction for a test instance can be obtained by running the Bayesian neural network multiple times and summing the $n$ samples. As a result, we obtain the output as a probability distribution for a test instance $(\hat{x}, \hat{y})$. Then, the prediction for that instance can be achieved by taking the mean of the output distribution, as shown in \eqref{eq4}. In addition, we can determine the uncertainty of the prediction for a test instance by calculating the variance of the output distribution, as shown in \eqref{eq5}:
\begin{equation}
p(\hat{y}|\hat{x}, \mathcal{D}_{tr}) = \int p(\hat{y}|\hat{x}, \theta)p(\theta, \mathcal{D}_{tr})\,d\theta \label{eq3}
\end{equation}
\begin{equation}
p(\hat{y}|\hat{x}, \mathcal{D}_{tr}) \approx \frac{1}{n}\sum_{i = 1}^{n} p(\hat{y}|\hat{x}, \theta) \label{eq4}
\end{equation}
\begin{equation}
v \approx \frac{1}{n}\sum_{i = 1}^{n} (p(\hat{y}|\hat{x}, \theta) - p(\hat{y}|\hat{x}, \mathcal{D}_{tr}))^2. \label{eq5}
\end{equation}
\begin{figure*}[!t]
\centering{\includegraphics[width=150mm, height = 36mm]{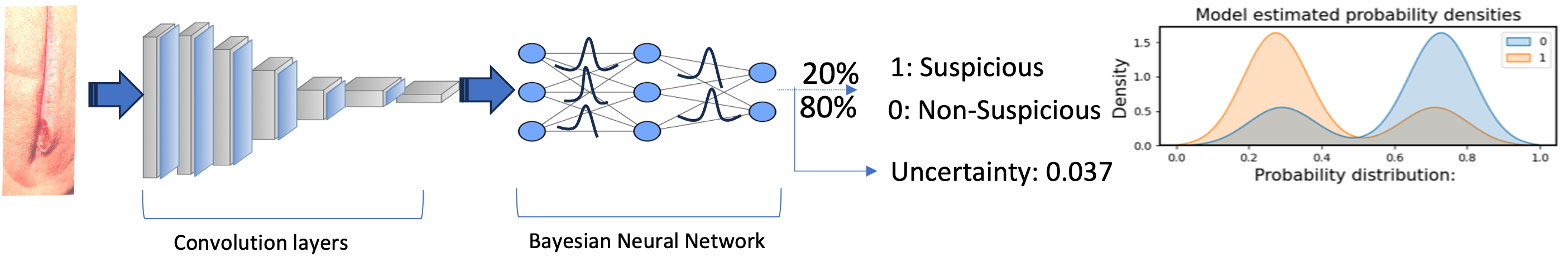}}
\caption{The architecture design for our hybrid Convolutional and Bayesian Neural Network.}
\label{fig1}
\end{figure*}

\begin{figure}[!t]
\centering{\includegraphics[width=\columnwidth]{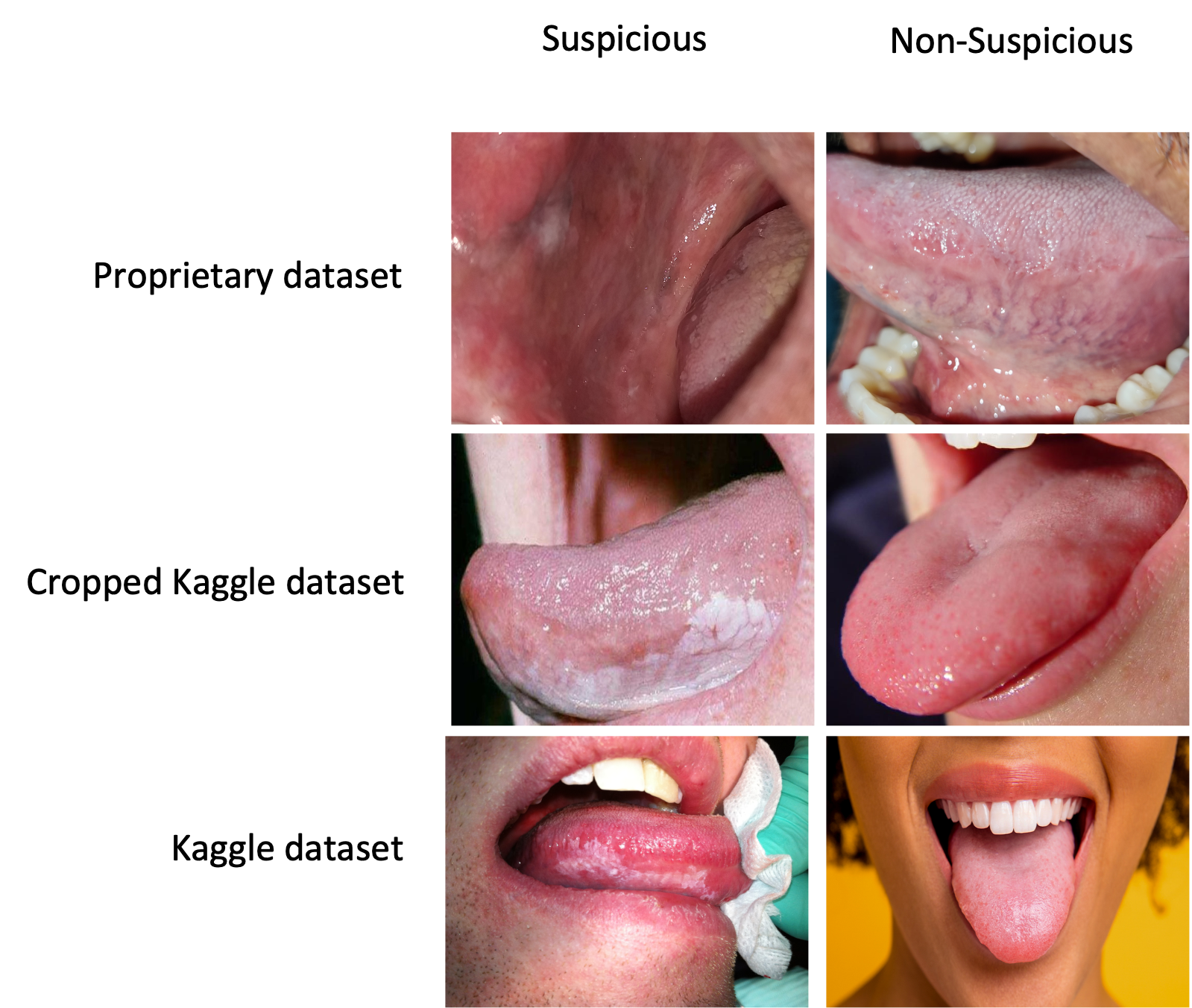}}
\caption{Sample images from the proprietary dataset, cropped Kaggle dataset, and Kaggle dataset.}
\label{fig1a}
\end{figure}

The proposed framework is illustrated in Fig. \ref{fig1}. Because training a complete Bayesian neural network is computationally expensive, we adopted a hybrid approach. The backbone of the proposed Bayesian neural network model is based on the MobileNet-v1 architecture, which comprises entirely of convolutional and depth-wise convolution layers. These layers take a high-dimensional input image (224x224x3) and extract the necessary features. At the head of the model, we introduced a dense variational layer in which we imposed a prior over the weights of that layer. For this study, we used a spike-and-slab prior and an output layer with a one-hot categorical dense variational layer from TensorFlow Probability, which generated a probability distribution of the output rather than a deterministic estimate. These layers operate as explained previously, providing predictions and associated uncertainties. The Bayesian neural network optimizes the objective function described in \eqref{eq2}. The KL divergence term in the objective function intuitively acts as a regularization term for the network, helping to reduce overfitting and improve generalizability for new data distributions, even with small amounts of data. The use of such Bayesian deep learning models in healthcare, where data availability is challenging, is extremely helpful. This is because, in addition to achieving accurate predictions, it is essential for a deep learning model to provide reliable uncertainty estimates to ensure its trustworthiness for clinicians and end users.

\section{Datasets and Preprocessing}
% \label{sec:guidelines}

In this study, we used two datasets: a proprietary oral cancer dataset collected by dentists in India \cite{b18} and an oral cancer dataset available on Kaggle \cite{b19}. In the proprietary dataset, all images were color images of different sizes captured from smartphones. The dataset contains 290 suspicious images and 551 non-suspicious images. These images were noisy, meaning each contained extra objects like gloves, tools, etc., and did not particularly focus on the regions where suspicious patches were present. These extra objects are referred to as noise. To reduce the computational requirement of searching for important features in the feature space, we cropped all extraneous images to focus on relevant areas. We then split the data into training and test sets by randomly picking 50 suspicious images and 50 non-suspicious images for the test dataset.

Because we used a MobileNet-v1 backbone-based Bayesian model, the available data were not sufficient to train this data-hungry model. To resolve this issue, we used augmentation methods such as rotation, shearing, scaling, etc., to increase the size of the dataset. These augmentation methods were selected to generate images that resemble images that users would take with their smartphones. After augmentation, we noticed that the algorithm generated some blank images and so we removed all outliers. Once the dataset was ready, we balanced it by down-sampling the major class to match the minor class. Finally, we used this balanced and augmented dataset of 3026 images, equally distributed between both the classes to train the Bayesian neural network. The test set was a standalone set and was not influenced by the pre-processing performed on the training set.

The second test dataset, was made up of oral cancer images available on the Kaggle platform. The dataset contains 131 color images, including 87 suspicious and 44 non-suspicious images. It comprises images of the lips and tongue, with and without cancerous patches. Since our proprietary training data primarily focused on the regions inside the mouth, we removed most of the lip images and retained only a few from the Kaggle dataset, treating these lip images as out-of-distribution data to evaluate our model’s reliability. This resulted in a dataset of 85 images, with 65 suspicious and 20 non-suspicious samples. We refer to this modified test set as the Kaggle dataset in this paper.

In this study, we used the Kaggle test data in two forms: with cropping and without cropping, to evaluate uncertainty quantification. For the second set, we selected samples that closely resembled our proprietary data and cropped out the extra noise. These images were cropped similar to our proprietary dataset, leaving the focus on the areas of interest. We refer to this second set as the Cropped Kaggle dataset, which contains 21 suspicious and 11 non-suspicious images. Sample images from the proprietary, Cropped Kaggle, and Kaggle datasets are shown in Fig. \ref{fig1a}.

\section{Results and Discussion}
In our experiments, we utilized the TensorFlow framework, incorporating the TensorFlow Probability module to develop a Bayesian variational layer comprising 32 nodes. To mitigate computational costs associated with training the entire model and to prevent overfitting due to limited data availability, we employed a transfer learning approach. This involved leveraging a pretrained model analogous to our target problem, followed by fine-tuning on the specific dataset. Consistent with our earlier methodology, we adopted the MobileNet-V1 architecture—excluding its classification head—as the backbone of our proposed Bayesian neural network. The pretrained weights from MobileNet-V1, trained on the ImageNet dataset, served as the initial weights for this backbone. Subsequently, we appended a variational layer and an output layer, as detailed in Section III. The variational layer was initialized with random parameters sampled from spike-and-slab priors.

We fine-tuned the Bayesian neural network using our training dataset to evaluate its uncertainty estimation capabilities and performance relative to classical deep neural networks. Advanced fine-tuning methods can be employed to identify the optimal model fit for a given dataset. For training, we utilized the RMSprop optimization algorithm with a learning rate of 0.01 and a batch size of 32. The model was trained over 150 epochs to minimize the negative log-likelihood loss function.

In addition, we developed a classical deep neural network to compare the performance of the proposed Bayesian model. In this classical neural network, we used the same mobilenet-v1 backbone with pre-trained weights for the ImageNet dataset that we used in our Bayesian deep neural network. Instead of a variational layer, we used a classical neural network layer with 32 nodes and an output layer with 1 node for binary classification. The initial weights for both the neural networks were randomly selected. We used the same parameters and transfer learning technique as before to run the classic model over 150 epochs.

A proprietary training dataset was used to train both models so that their performances could be compared fairly. The experiments were performed using the high-performance computing resources available at the University of Texas at Dallas. The optimized model was obtained by preserving the best weights. After training the models, we evaluated them on our proprietary test dataset (kept separate from the training dataset). In addition, we evaluated the proposed model on the Kaggle dataset to evaluate its generalizability. During the test, we predicted an input using the method described in \eqref{eq3} by running the Bayesian model 50 times (here $n = 50$).

We found that our Bayesian deep neural network performed equivalently well compared to the classical deep neural network on proprietary test data, achieving 94\% and 94.99\% accuracy, respectively, as shown in Table I. We also investigated the generalization capabilities of both models on the Kaggle dataset with and without noise. With noise, the Bayesian model achieved 88\% accuracy, which is significantly better than the classic model (72.94\% accuracy). We hypothesize that the presence of noisy data increased the difficulty of feature extraction and image classification. On the cropped Kaggle dataset, both models performed well; however, the Bayesian model achieved 90\% accuracy, surpassing the classical model's 87\% accuracy. The results demonstrate that with a small dataset, the classical model generalized well on data with the same distribution as the training dataset but failed to generalize on data with different distributions. In contrast, the Bayesian model generalized much better even with noisy datasets. This is because the Bayesian model learned the ability to properly extract important features and classify inputs from the noisy feature space, even when trained on a small dataset, which is often the likely scenario in the medical domain.

\begin{table}
\caption{
Results
}
\setlength{\tabcolsep}{3pt}
\begin{tabular}{|p{65pt}|p{25pt}|p{25pt}|p{45pt}|p{45pt}|}
\hline
Models&
Training Data&
Test Data&
Kaggle Test Data&
Cropped Kaggle Data \\
\hline
CNN-Bayesian&
99.99\%&
94\% &
88\% &
90\% \\

CNN&
99.99\% &
94.99\% &
72.94\% &
87\% \\
\hline
\end{tabular}
\label{tab1}
\end{table}

\begin{figure}[!t]
\centering{\includegraphics[width=\columnwidth]{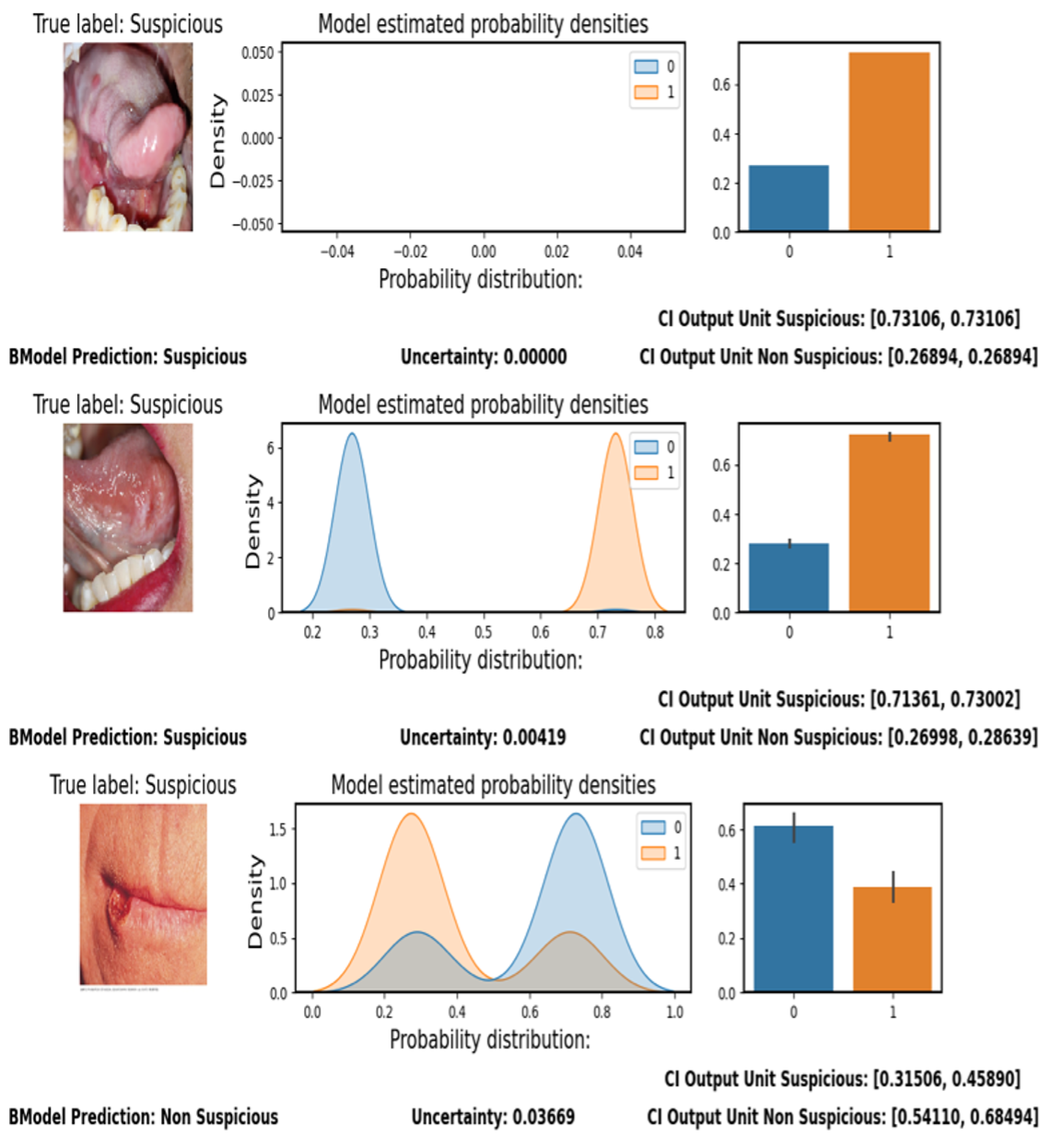}}
\caption{Performance analyses plot for our Bayesian model using uncertainty quantification.}
\label{fig2}
\end{figure}

In addition to predicting the target output, the proposed Bayesian neural network also estimates the uncertainty associated with its weight parameters for each prediction. This enables us to interpret the model's confidence and, consequently, determine whether its prediction can be trusted. Such interpretability is particularly critical in high-risk domains like healthcare. One practical application of this capability is to assess the quantified model uncertainty and decide whether to accept a prediction or refer it for further evaluation. In healthcare settings, we expect the model to provide outputs with near 100\% confidence (i.e., low uncertainty). If the model exhibits low uncertainty, the prediction may be considered reliable. However, when the uncertainty is high, the model's confidence is low, and the prediction should be treated with caution. In such cases, the image can be flagged for review by a medical specialist based on a confidence threshold—for instance, 99\%. These advanced computations can be integrated into an automated diagnostic system, enabling rapid screening and improving patient care in resource-limited environments.

Taking this into account, we used the uncertainty output from our Bayesian model to produce the plots shown in Fig. \ref{fig2}. Each row in the figure corresponds to a different test image. In each plot, the test image is displayed on the left, with the model’s prediction shown at the bottom left. For verification, the ground truth (true label) is shown at the top left, above the image. In the center of the plot, we included a kernel density estimate (KDE) graph to visualize the uncertainty through the probability distribution of each output class (0: Non-Suspicious, 1: Suspicious). This distribution was derived by sampling the model's weights 50 times. The KDE plot provides a continuous representation of how the model’s predictions are distributed across the probability range.

The quantified output uncertainty value is shown at the bottom center of the plot. On the right side, we included a bar graph representing the softmax activation values for each output class. Additionally, we displayed the 95\% confidence intervals for each class at the bottom right corner of the plot, with corresponding indicators (black lines) overlaid on the bar graph. These confidence intervals represent the range within which we are 95\% confident that the true probability lies. They were computed by running the Bayesian neural network with 50 weight samples to approximate the posterior distribution, as described in \eqref{eq3}.

 The Bayesian model, as shown in Fig. \ref{fig2}, is 100\% confident in its prediction that the first test image is suspicious. There are no continuous probability distributions or error bars in the corresponding bar graph. For the second test image (middle plot), the model displays continuous probability distributions for two output units with slight overlap, indicating minor uncertainty. The uncertainty value for the correct prediction of the suspicious input is 0.00419. This slight under-confidence can also be verified by analyzing the confidence intervals, which are influenced by the spread and overlap of the two output distributions. The peaks of these distributions indicate where the model's predictions were most concentrated for each class.

For the third test image, we used a Kaggle test image showing lips—a distribution unfamiliar to the Bayesian model based on its training data. Consequently, we expect a confused prediction. The model outputs spread-out continuous probability distributions for each class, with a high uncertainty value of 0.03669. These broad distributions result in wide confidence intervals: [0.31506, 0.45890] for the suspicious class and [0.54110, 0.68494] for the non-suspicious class. Moreover, the overlap between the distributions reduces the gap between the two classes, increasing the model’s confusion and reducing its classification certainty. Overall, due to ambiguous features and high uncertainty, the model struggled to make a reliable decision and misclassified the test image.

\begin{figure}[!t]
\centering{\includegraphics[width=\columnwidth]{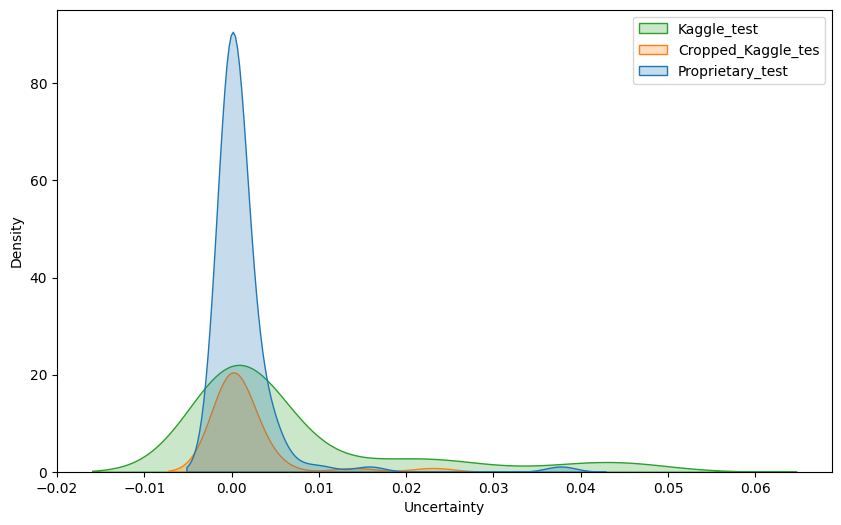}}
\caption{Kernel density estimate (KDE) plot showing the relationship between the prediction distribution and the associated uncertainties for different test datasets. The plot shows how uncertainties vary with the prediction distributions for the datasets under evaluation.}
\label{fig3}
\end{figure}

We aimed to evaluate the uncertainty estimation capability of the proposed Bayesian deep learning model. A well-calibrated Bayesian model is expected to exhibit high uncertainty when presented with data distributions that differ from those seen during training. This is because unfamiliar distributions often contain features the model has not previously encountered, and a high uncertainty value should reflect this lack of knowledge. For instance, in our case, the Bayesian model was trained exclusively on data centered around the region of interest (i.e., cancerous patches). Therefore, when the model encounters unfamiliar elements—such as gloves, which it was not trained to distinguish from cancerous tissue—it should indicate a higher level of uncertainty. This behavior enables medical practitioners and users to assess the model's confidence, thereby fostering greater trust in its predictions.

Figure \ref{fig3} presents the kernel density estimate (KDE) plot, illustrating the relationship between prediction uncertainty and its distribution across different test datasets. The x-axis denotes the uncertainty values, while the y-axis represents the density of these values among the test instances. The results show that the proposed Bayesian deep neural network is highly confident in its predictions on the proprietary test dataset, which closely matches the training data distribution. Similarly, the model maintains high confidence when predicting on the cropped Kaggle dataset, as the cropping aligns its distribution more closely with that of the proprietary training data. In contrast, the model exhibits significantly higher uncertainty on the uncropped Kaggle dataset, as reflected by the broader spread of uncertainty values in the KDE plot. These findings demonstrate that the proposed Bayesian model provides well-calibrated uncertainty estimates. This capability is especially critical in healthcare, where practitioners routinely navigate uncertainty. A deep learning model that mimics this cautious behavior can meaningfully support and enhance clinical decision-making.

\begin{figure*}[!t]
\centering{\includegraphics[width=160mm]{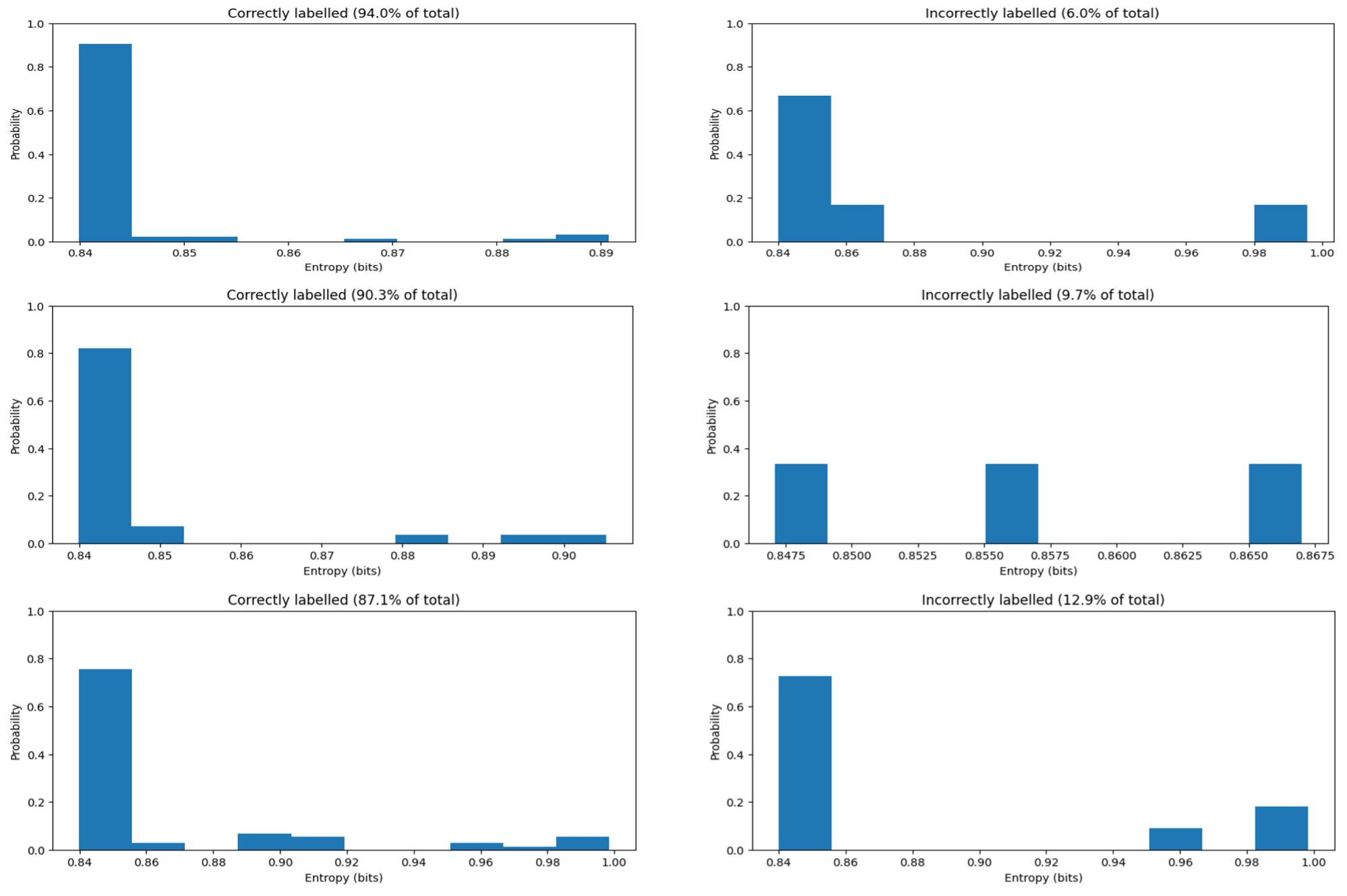}}
\caption{Uncertainty measurement plots for each test dataset, evaluating the uncertainty for correctly and incorrectly classified samples. The figure displays three plots arranged vertically: the top plot shows the uncertainty measurements for the proprietary test dataset, the middle plot shows the cropped Kaggle test dataset, and the bottom plot shows the Kaggle dataset.}
\label{fig4}
\end{figure*}

We further investigated the uncertainty estimates of our Bayesian deep neural network on both correctly and incorrectly labeled samples across each test dataset to gain a detailed understanding of the model's uncertainty behavior. A well-calibrated Bayesian model is expected to exhibit high uncertainty for misclassified instances, as such errors often arise when the model encounters features not previously seen during training. These misjudgments reflect a knowledge gap, which should intuitively correspond to elevated uncertainty values. This behavior is especially critical in high-risk domains such as healthcare, where diagnostic models must not only deliver accurate predictions but also signal under-confidence when uncertain.

In contrast, conventional deep neural networks typically fail to capture this nuance—they often remain overconfident in their predictions, even when operating outside their learned distribution. Such overconfidence can be dangerous in real-world diagnostic applications. Hence, incorporating uncertainty estimation into deep learning models is essential for safe and trustworthy deployment in healthcare settings.

To evaluate the model’s uncertainty behavior, we employed an uncertainty distribution graph. Figure \ref{fig4} displays the distribution of entropy bins, which quantify uncertainty, for correctly classified (left) and incorrectly classified (right) instances produced by the Bayesian deep neural network across the test datasets. The x-axis represents entropy values (a proxy for uncertainty), and the y-axis shows the fraction of instances falling into each bin. As expected, predictions with low entropy (toward the left side of the plot) are predominantly correct, indicating that the model’s confidence is well-aligned with its performance. In contrast, misclassified instances are more concentrated in higher-entropy bins, reinforcing the model’s ability to express appropriate under-confidence when making incorrect predictions.

From Fig. \ref{fig4}, we observe that our Bayesian neural network correctly classified 94\%, 90\%, and 87\% of the total samples, and misclassified 6\%, 9.7\%, and 12.9\% of the total samples from the proprietary, Kaggle, and cropped Kaggle test datasets, respectively. The results further demonstrate that the majority of correctly classified samples exhibit low entropy values (approximately 0.84–0.85 bits), indicating high model confidence and low uncertainty. In contrast, the entropy distribution for the misclassified samples is more dispersed, reflecting the model’s higher uncertainty in cases of incorrect predictions.

Theoretically, a Bayesian neural network should also display high uncertainty when encountering data from distributions that differ significantly from the training data. Our experiments confirm this behavior, as seen with the noisy Kaggle dataset. The entropy distribution for misclassified samples in the middle graph shows an even spread, underscoring the model’s uncertainty. Additionally, the two peaks on the far right of the bottom graph represent high entropy values for incorrectly labeled samples from the cropped Kaggle dataset, which—despite resembling the training distribution—still introduced unfamiliar variations to the model.

It is also important to note that the model exhibited overconfidence in a subset of misclassified samples from both the proprietary and cropped Kaggle datasets. This is evident from the small peaks in sample density on the left side of the top and bottom graphs, indicating low entropy values despite incorrect predictions. This suggests potential areas for improvement that could be addressed by augmenting the training dataset with more diverse examples.

Moreover, several peaks appear on the right side of the correctly labeled section, implying that some correctly classified samples still carry high uncertainty. This reflects the model's recognition of its knowledge limitations—even when producing accurate predictions. This phenomenon is particularly evident in the Kaggle datasets, where certain images contain features such as lips, which were not part of the training data. To address this, we can analyze these high-uncertainty yet correctly classified samples and enhance the model by incorporating similar data into the training set.

Finally, the model’s uncertainty output can be leveraged to flag high-risk cases for further expert review. By referring predictions with high uncertainty for specialist evaluation, we can improve both the reliability and clinical acceptance of the model’s results—an essential consideration in medical applications.

\section{Conclusion}
In conclusion, this study underscores the critical importance of early detection in combating oral cancer and highlights the potential of deep learning to address this pressing healthcare challenge. Nevertheless, the reliability of deep learning models remains a major concern, particularly in real-world clinical settings where data is often limited and variable. To address this, we proposed a hybrid Convolutional and Bayesian neural network architecture that enhances model reliability through uncertainty quantification.

Our results demonstrate that the proposed model can effectively classify real-world oral images, even when trained on limited data. The model exhibited strong generalization capabilities, achieving notable accuracy on the Kaggle dataset, including in the presence of noise. Most importantly, the Bayesian model consistently produced low uncertainty for correctly classified samples and high uncertainty for misclassified ones—indicating calibrated confidence in its predictions, a critical requirement for healthcare applications.

We believe this work advocates for the broader adoption of Bayesian deep learning in the medical domain. As a next step, we plan to conduct extensive field validation and integrate the proposed model into a practical AI framework for reliable and accessible oral cancer screening.

\section*{Acknowledgment}
The authors thank Dr. C.S. Mani (dr.cs.mani@gmail.com) and Lakshmi Narayana
(drnarayana777@gmail.com) of the Cancer Research and Relief Trust, Chennai, India, for their assistance in collecting and providing the proprietary oral cancer dataset.

\EOD

\end{document}